\pdfoutput=1

\documentclass[11pt]{article}

\usepackage[table]{xcolor}
\usepackage{acl}

\usepackage{graphics}
\usepackage{booktabs}
\usepackage{times}
\usepackage{latexsym}
\usepackage[T1]{fontenc}
\usepackage[utf8]{inputenc}
\usepackage{microtype}
\usepackage{hyperref}
\usepackage{longtable}
\usepackage{multirow}
\usepackage{graphicx}
\usepackage{float}
\usepackage{cleveref}
\usepackage{balance}

\usepackage{tabularx,colortbl}

\usepackage{tikz}
\usepackage{collcell}

\def\hyphenateAndTtWholeString #1{\xHyphenate#1$\wholeString\unskip}

\def\xHyphenate#1#2\wholeString {\if#1$%
    \else\transform{#1}%
    \takeTheRest#2\ofTheString\fi}

\def\takeTheRest#1\ofTheString\fi
{\fi \xHyphenate#1\wholeString}

\def\transform#1{\url{#1}\hskip 0pt plus 1pt}

\def\urlx #1{\href{#1}{\hyphenateAndTtWholeString{#1}}}

\usepackage{etoolbox}

\newtoggle{inTableHeader}
\toggletrue{inTableHeader}
\newcommand*{\StartTableHeader}{\global\toggletrue{inTableHeader}}%
\let\OldTabular\tabular%
\let\OldEndTabular\endtabular%
\renewenvironment{tabular}{\StartTableHeader\OldTabular}{\OldEndTabular\StartTableHeader}%

\newcommand*{\MinNumber}{0.0}%
\newcommand*{\MidNumber}{30.0} %
\newcommand*{\MaxNumber}{100.0}%

\newcommand{\ApplyGradient}[1]{%
  \iftoggle{inTableHeader}{#1}{
    \ifdim #1 pt > \MidNumber pt
        \pgfmathsetmacro{\PercentColor}{max(min(100.0*(#1 - \MidNumber)/(\MaxNumber-\MidNumber),100.0),0.00)} %
        \hspace{-0.33em}\colorbox{yellow!\PercentColor!blue}{#1}
    \else
        \pgfmathsetmacro{\PercentColor}{max(min(100.0*(\MidNumber - #1)/(\MidNumber-\MinNumber),100.0),0.00)} %
        \hspace{-0.33em}\colorbox{blue!\PercentColor!blue}{#1}
    \fi
  }}
\newcolumntype{R}{>{\collectcell\ApplyGradient}c<{\endcollectcell}}

\title{Building Extractive Question Answering System to Support \\ Human-AI Health Coaching Model for Sleep Domain}

\author{Iva Bojic$^{1}$ \and Qi Chwen Ong$^{1}$ \and Shafiq Joty$^{2,3}$ \and Josip Car$^{1,4}$ \\
$^1${Centre for Population Health Sciences, Lee Kong Chian School of Medicine, NTU, Singapore} \\
$^1${School of Computer Science and Engineering, NTU, Singapore} \\
$^3${Salesforce AI Research, USA} \\
$^4${Department of Primary Care and Public Health, School of Public Health, Imperial College London, UK} \\
}

\begin{document}
\maketitle

\section{Introduction}

Non-communicable diseases are the leading cause of deaths globally, imposing a huge burden on healthcare systems \cite{world2014global}. This leads to the increasing emphasis on primary prevention, which seeks to avert the onset of diseases by addressing their causes and risk factors, such as altering unhealthy behaviors \citep{kisling2021prevention}. Health coaching has emerged as an effective strategy for primary prevention of chronic conditions by facilitating lifestyle behavior change \cite{yang2020current}. With assistance from Question Answering (QA) system, health coaches could be empowered with evidence-based information, streamlining the task of helping clients to adopt new practices.

\begin{figure}[ht!]
\centering
\includegraphics[width=1\columnwidth]{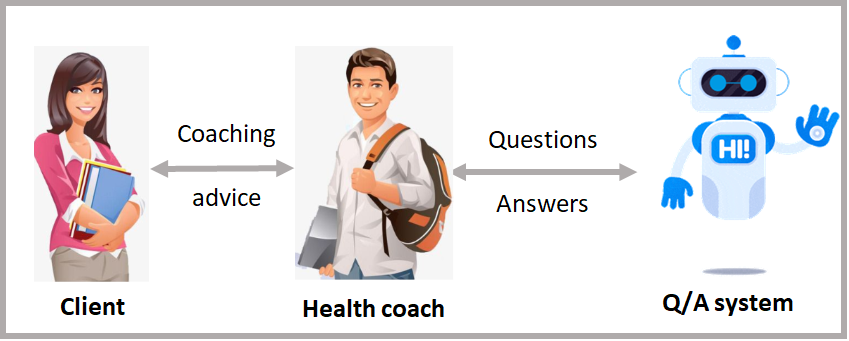}
\caption{Human-AI health coaching model.}
\label{fig1}
\end{figure}

An innovative human-Artificial Intelligence (AI) health coaching model incorporating QA systems holds the potential to revolutionize the health coaching domain and the practice of preventive healthcare. This model involves the interaction of three components: a health coach, a client, and a QA system (see Fig. \ref{fig1}). The coaching process takes place online (e.g., through text messaging). If needed, the coach can consult the QA system for assistance in responding to \textit{factoid} questions posed by the client \cite{wang2006survey}. In this way, human coaching is supported by the evidence-based knowledge supplied by the QA system. This not only accelerates the health coaches' response time but also ensures that the information provided by the health coach is grounded in evidence.

This paper outlines the creation of a domain-specific extractive QA system employed as the AI component within a Human-AI health coaching model. Generally, an extractive QA system composes two phases: a passage \textit{retrieval} followed by a passage \textit{reader} \citep{chen2017reading}. For a given question, the retrieval initially extracts a set of pertinent passages from a knowledge base (e.g., a text corpus), after which the reader selects an answer (for instance, a text span) from one of those passages \citep{zhu2021retrieving}. Although generative large language models like ChatGPT show great ability in QA, the lack of references and hallucinations limit their utilizations in healthcare at present \cite{zhang2023language}. \emph{Domain-specific} systems typically address questions confined to a single area of expertise, such as medicine \citep{kia2022adaptable}.

\section{Methods}

In the pursuit of developing a domain-specific extractive QA system to bolster a Human-AI health coaching model, we manually assembled a dataset called \textit{SleepQA}, comprising 7,005 passages focused on the sleep domain \cite{bojic2022sleepqa}. This dataset featured 5,000 expert-annotated labels\footnote{Code and dataset are available at: \\ \url{https://github.com/IvaBojic/SleepQA}}. Subsequently, we fine-tuned various domain-specific \textit{BERT} models on our \textit{SleepQA} dataset and conducted a comprehensive assessment of the resulting end-to-end extractive QA system using both automatic and human evaluation methods. 

To further improve the QA system's performance, we introduced a data-centric framework aimed at refining domain-specific extractive QA datasets\footnote{Code and datasets are available at: \\ \url{https://github.com/IvaBojic/framework}}. Our proposed framework consists of methods for (i) improving the selection of negative passages during the retrieval fine-tuning, and (ii) question reformulation through paraphrasing, word replacement, and back translation \cite{bojic2023data}. Using the proposed framework, we first produced new datasets by employing the four aforementioned data quality enhancement methods. Subsequently, we fine-tuned both the retrieval and reader models individually on each new dataset. Following this, we fine-tuned the retrieval/reader models continuously, commencing from the best individual checkpoint, using the enhanced datasets that demonstrated progress in the initial step. Lastly, we generated new augmented datasets by merging all datasets that exhibited fine-tuning enhancements in the initial step.

\section{Results}

The initial fine-tuned QA system consisted of fine-tuned \textit{PubMedBERT}\footnote{\urlx{https://huggingface.co/microsoft/BiomedNLP-PubMedBERT-base-uncased-abstract-fulltext}} retrieval model and \textit{BioBERT BioASQ}\footnote{\url{https://huggingface.co/gdario/biobert_bioasq}} reader model. Automatic evaluation of the initial fine-tuned QA system on test labels demonstrated that the proportion of accurate predictions, specifically those that precisely align with the correct answers (known as an Exact Match or EM), reached 24\%, as opposed to the 30\% achieved by the baseline QA system (see Table \ref{tab1}). The baseline QA system refers to the combination of \textit{Lucene BM25} \citep{Lin_etal_SIGIR2021_Pyserini} used as a baseline retrieval model and \textit{BERT SQuAD2 QA} \citep{rajpurkar2018know} as a baseline reader model.

 \begin{table}[ht!]
 \centering
 \caption{Automatic evaluation of two extractive QA systems on test labels using EM \cite{bojic2022sleepqa}.}
 \begin{tabular}{l|rr}
 \toprule
 \textbf{Extractive QA system name} & \textbf{EM} \\
 \midrule 
 \textit{Lucene BM25} + \textit{BERT SQuAD2} QA & 0.30 \\
 \textit{PubMedBERT} + \textit{BioBERT BioASQ} & 0.24 \\
 \bottomrule
 \end{tabular}
 \label{tab1}
\end{table}

Although our initial fine-tuned QA system did not outperform the baseline on the test labels, results from human evaluation of the two QA systems on real-world questions suggested otherwise. Real-world questions are sleep-related questions posed by university students, which mimic the types of questions that clients would ask their sleep health coaches in real life. The findings presented in \cite{bojic2022sleepqa} showed that our initial fine-tuned QA system outperformed the baseline QA system in almost 40\% of cases in its intended purpose, which is accurately providing sleep health coaches with the appropriate responses to sleep-related inquiries they might receive from their clients.

In order to increase the performance of our initial fine-tuned QA system on the test labels as well, we introduced various data quality enhancement methods. The best method to increase the performance of retrieval fine-tuning was \textit{word substitution} while fine-tuning of the reader model resulted in the best results when using \textit{augmentation}. The overall relative increase in performance of the entire enhanced fine-tuned QA system, compared to the initial one, was 17\% (i.e. EM = 0.28). Although the relative increase was not negligible, the enhanced fine-tuned QA system still did not outperform the baseline QA system on the test labels.

\section{Discussion and Conclusions}

This paper outlines the creation of a domain-specific extractive QA system, tailored for the sleep domain, utilizing a two-stage approach. First, we fine-tuned our initial QA system using 5,000 experts-annotated labels from the \textit{SleepQA} dataset. Second, we enhanced the initial \textit{SleepQA} dataset by employing a data-centric approach and subsequently fine-tuned the QA system on the enhanced dataset. The enhanced QA system was then integrated into a Human-AI health coaching model.

Automatic evaluation showed that both fine-tuned QA systems did not surpass the performance of the baseline QA system. Despite employing data-centric methods to reformulate manually collected questions, it is likely that the similarities remain between initial and reformulated questions. The superiority of the \textit{Lucene BM25} approach over all fine-tuned retrievals could be attributed to this.

Nevertheless, human evaluation indicated that the initial fine-tuned QA system outperformed the baseline, displaying significant improvement in domain-specific natural language processing on real-world sleep-related questions. Human evaluation of enhanced fine-tuned QA system was done during a pilot Randomized Controlled Trial (RCT). Participants in the intervention arm interacted with health coaches who were supported by the enhanced fine-tuned QA system, while in the control group, they directly interacted with the QA system without the intermediate interaction with a coach. Our future work includes analyzing the 5,000 questions and participants' feedback on the provided answers collected from this pilot RCT.

\section*{Acknowledgment}

The authors would like to acknowledge the Accelerating Creativity and Excellence (ACE) Award (NTU-ACE2020-05) and center funding from Nanyang Technological University, Singapore. Josip Car’s post at Imperial College London is supported by the NIHR NW London Applied Research Collaboration.

\balance
\bibliography{references}
\bibliographystyle{acl_natbib}

\appendix

\end{document}